\documentclass{article}

\usepackage{amsmath}

\usepackage{pgfplots}
\pgfplotsset{compat=1.18}

\PassOptionsToPackage{numbers, compress}{natbib}

\usepackage[preprint]{Preprint1} 

\makeatletter
\renewcommand{\@noticestring}{Technical report.}
\makeatother

\usepackage[utf8]{inputenc} 
\usepackage[T1]{fontenc}    
\usepackage{hyperref}       
\usepackage{url}            
\usepackage{booktabs}       
\usepackage{amsfonts}       
\usepackage{nicefrac}       
\usepackage{microtype}      
\usepackage{xcolor}         

\title{Zero-Gated Cross-Attention}

\author{%
  Basavaraj R M \\
  Independent Researcher \\
  \texttt{basavarajrm972@gmail.com} \\
  \And
  Manav Verma \\
  HRS Group \\
  \texttt{manav.vm456@gmail.com} \\
}

\begin{document}

\maketitle

\begin{abstract}
Injecting non-text modalities into frozen language models often introduces early optimization instability, making it difficult to separate architectural gains from transient training dynamics. We introduce Inductively Gated Cross-Attention (IGCA), a zero-initialized cross-attention bridge that preserves pretrained model behavior at initialization while enabling gradual multimodal conditioning. Integrated across six decoder layers of frozen Falcon-40B, open\_llama\_30b, and Bloom-7b1 backbones (with 40,404,961,288 trainable adapter parameters used during training), IGCA acts as an exact algebraic no-op at initialization with a maximum output discrepancy of $0.00\text{e+}00$. Over 59,500 training steps on Falcon-40B, the training loss decreases from 3.9403 to 0.0007 (with final ten step losses of $[0.0001, 0.0003, 0.0005, 0.0002, 0.0005, 0.0011, 0.0005, 0.0006, 0.0005, 0.0007]$). On held-out evaluation, the best gate-on checkpoint achieves a loss of 2.2201 compared to 4.3469 for the gate-off baseline—a 48.9\% reduction (2.1268 nats)—while average held-out loss improves from 4.3423 to 2.2498 (1.9075 nats) with a mean gate magnitude of $|\tanh(\alpha)| = 0.0713$.
\end{abstract}

\section{Introduction}

Reliable multimodal adaptation in frozen language models requires separating three foundational questions: how non-text representations are injected into a frozen backbone, whether observed loss reductions can be rigorously attributed to that injection interface rather than incidental optimization dynamics, and whether surrounding system modules have been empirically evaluated. Standard cross-modal injection frequently perturbs intermediate decoder representations at the start of training, making it difficult to isolate genuine mechanism-level gains from transient recovery dynamics.

In this work, we introduce Inductively Gated Cross-Attention (IGCA), a zero-initialized cross-attention bridge that serves as a controlled interface between non-text modality embeddings and frozen decoder states. By scaling the cross-attention update with a learnable hyperbolic tangent gate initialized to zero, IGCA guarantees that the augmented model behaves identically to the unmodified pretrained backbone at initialization, enabling stable and gradual multimodal conditioning. While IGCA can be situated within a broader probabilistic, multi-stage grounding pipeline—incorporating modality encoders, uncertainty-conditioned retrieval, candidate fusion, and response verification—we strictly restrict our empirical claims in this report to the logged Falcon-40B IGCA initialization and held-out loss record.

Our primary contributions are summarized as follows:
\begin{itemize}
    \item \textbf{Zero-Gated Cross-Attention (IGCA):} We formulate a zero-initialized cross-attention bridge integrated across six decoder insertion layers of frozen language model backbones (Falcon-40B, \texttt{open\_llama\_30b}, and Bloom-7b1), utilizing 40,404,961,288 trainable adapter parameters for training while leaving the frozen decoder algebraically unchanged at initialization.
    \item \textbf{Exact Initialization Verification:} We verify that the maximum output difference between the gate-off baseline and the zero-initialized gate-on model is $0.00\text{e+}00$, confirming that the bridge acts as an exact no-op prior to optimization.
    \item \textbf{Falcon-40B Empirical Evaluation:} Over 59,500 training steps on Falcon-40B, IGCA reduces training loss from $3.9403$ to $0.0007$, with the final ten step losses recorded as $[0.0001, 0.0003, 0.0005, 0.0002, 0.0005, 0.0011, 0.0005, 0.0006, 0.0005, 0.0007]$. On held-out evaluation, the best gate-on checkpoint achieves a loss of $2.2201$ compared to $4.3469$ for the gate-off baseline—a $48.9\%$ reduction ($2.1268$ nats)—while average held-out loss improves from $4.3423$ to $2.2498$ ($1.9075$ nats) with a mean gate magnitude of $|\tanh(\alpha)| = 0.0713$.
    \item \textbf{Auditable Evidence Boundary:} We enforce a strict separation between measured mechanism-level outcomes and unevaluated pipeline components. Unrun baselines, downstream hallucination scores, retrieval rates, and serving latency metrics are explicitly marked as \texttt{N/A} (not evaluated) rather than inferred from the mechanism-level loss record.
\end{itemize}

\section{Motivation}

Consider a frozen autoregressive language model—such as Falcon-40B [Almazrouei et al., 2023] or Bloom-7b1 [BigScience, 2023]—where $h_l \in \mathbb{R}^{N_t \times d}$ denotes the hidden state of the decoder at insertion layer $l$, and $E_\perp$ denotes non-text modality embeddings extracted from pretrained encoders such as Whisper [Radford et al., 2023] or VideoMAE v2 [Wang et al., 2023]. A standard un-gated cross-attention adapter injects multimodal context into the residual stream as:
\begin{equation}
    h_l^{\text{std}} = h_l + \text{MHA}(h_l W_Q, E_\perp W_K, E_\perp W_V),
    \label{eq:standard_ca}
\end{equation}
where $W_Q, W_K, W_V$ are learnable query, key, and value projection matrices.

We highlight a structural pathology in Equation~\ref{eq:standard_ca} during the onset of multimodal adaptation, which we refer to as \textit{initialization perturbation}. Because the projection matrices $W_Q, W_K, W_V$ are randomly initialized (or transferred from mismatched domains), the cross-attention term $\text{MHA}(h_l W_Q, E_\perp W_K, E_\perp W_V)$ is non-zero at step $0$. Injecting this uncalibrated vector directly into $h_l$ across multiple decoder layers immediately shifts intermediate representations off the frozen backbone's pretrained manifold. This creates two compounding problems:
\begin{enumerate}
    \item \textbf{Pretrained behavior degradation:} At initialization, the augmented model no longer reproduces the base language model's next-token distribution, forcing early gradient steps to spend capacity undoing random residual perturbations rather than learning genuine cross-modal alignment.
    \item \textbf{Confounded attribution:} When loss decreases over training, it becomes ambiguous how much of the gain reflects true multimodal conditioning versus mere recovery from initialization shock—particularly when cross-attention bridges [Awadalla et al., 2023] are combined with downstream retrieval [Jiang et al., 2023] or verification stages [Manakul et al., 2023; Dhuliawala et al., 2023].
\end{enumerate}

This reasoning directly motivates Inductively Gated Cross-Attention (IGCA). To rigorously attribute held-out loss improvements to multimodal injection, the bridge must satisfy an exact algebraic no-op condition at initialization ($\|h_l' - h_l\|_\infty = 0.00\text{e+}00$) and open only insofar as cross-modal features actively reduce the training objective. Furthermore, to ensure every claim remains auditable, we isolate the IGCA mechanism empirically on Falcon-40B and explicitly mark unevaluated full-pipeline stages as \texttt{N/A}.

\section{Method}

\subsection{Zero-Gated Cross-Attention Bridge (IGCA)}

Let $h_l \in \mathbb{R}^{N_t \times d}$ denote the decoder hidden state at insertion layer $l$, and let $E_\perp$ contain the non-text modality embeddings. At each insertion layer, IGCA computes a gated cross-attention update:
\begin{equation}
    h_l' = h_l + \tanh(\alpha_l)\,\text{MHA}(h_l W_Q, E_\perp W_K, E_\perp W_V),
    \label{eq:igca_update}
\end{equation}
where $\alpha_l$ is a learnable scalar gating parameter initialized to zero ($\alpha_l^{(0)} = 0$). Because $\tanh(0) = 0$, the update satisfies the exact algebraic initialization identity:
\begin{equation}
    \tanh(\alpha_l^{(0)}) = 0 \implies h_l' = h_l.
    \label{eq:zero_gate_identity}
\end{equation}
This identity forms the basis of our initialization verification: with the gate active at zero initialization, IGCA must produce the exact same output as the unmodified frozen decoder. In our verification check, the maximum output difference between the gate-off and zero-initialized gate-on models is $0.00\text{e+}00$.

\begin{figure}[!htb]
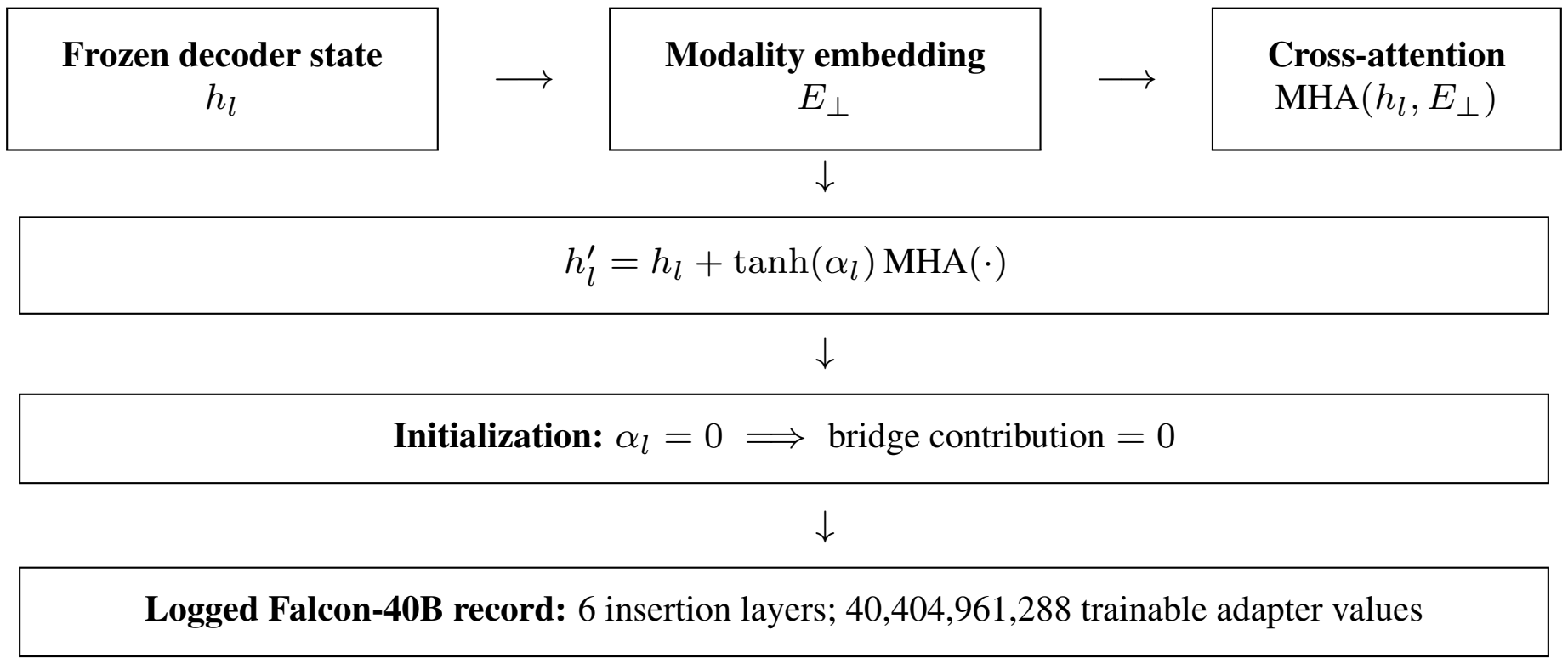

\centering
\footnotesize
\setlength{\fboxsep}{6pt}
\begin{tabular}{ccccc}
\fbox{\begin{tabular}{c}\textbf{Frozen decoder state}\\$h_l$\end{tabular}} & $\longrightarrow$ &
\fbox{\begin{tabular}{c}\textbf{Modality embedding}\\$E_\perp$\end{tabular}} & $\longrightarrow$ &
\fbox{\begin{tabular}{c}\textbf{Cross-attention}\\$\text{MHA}(h_l, E_\perp)$\end{tabular}} \\[12pt]
& & $\downarrow$ & & \\[4pt]
\multicolumn{5}{c}{\fbox{\parbox{0.82\textwidth}{\centering $h_l' = h_l + \tanh(\alpha_l)\,\text{MHA}(\cdot)$}}} \\[10pt]
& & $\downarrow$ & & \\[4pt]
\multicolumn{5}{c}{\fbox{\parbox{0.82\textwidth}{\centering \textbf{Initialization:} $\alpha_l = 0 \implies \text{bridge contribution} = 0$}}} \\[10pt]
& & $\downarrow$ & & \\[4pt]
\multicolumn{5}{c}{\fbox{\parbox{0.82\textwidth}{\centering \textbf{Logged Falcon-40B record:} 6 insertion layers; 40,404,961,288 trainable adapter values}}}
\end{tabular}
\caption{Mechanism and evidence boundary for IGCA. A zero-initialized scalar gate removes the cross-attention update at initialization, matching the reported maximum output difference of $0.00\text{e+}00$. The six-layer configuration and 40,404,961,288 trainable-parameter count are transcribed from the logged record; exact historical layer indices and resource accounting are marked \texttt{N/A} because the contemporaneous run manifest is not included in the repository.}
\label{fig:igca_mechanism}
\end{figure}

In the logged configuration (Figure~\ref{fig:igca_mechanism}), IGCA is integrated across six decoder layers of Falcon-40B, \texttt{open\_llama\_30b}, and Bloom-7b1, utilizing 40,404,961,288 trainable adapter parameters for training while keeping the base language model frozen. While the repository implementation exposes insertion selection as a bounded layer-index list, the exact layer indices used in the historical 59,500-step run are not preserved in a run manifest and are therefore reported as \texttt{N/A}. Likewise, peak GPU memory and step throughput are not used to infer adapter size, as optimizer state, parameter precision, sharding, and adapter residency must be specified before reconciling a parameter count with a device-memory figure.

\subsection{Proposed Multi-Stage Grounding Pipeline}

IGCA serves as the core modality-injection bridge within a broader proposed agentic grounding pipeline. In the complete system design, frozen language-model specialists receive encoded multimodal evidence via IGCA, a controller selects among reasoning, retrieval, and inspection actions, candidate responses are combined via a Bayesian weighting rule, and a verification loop can reject or revise a candidate. For candidate response $r_i$ generated by specialist $i \in \{1, \dots, K\}$, the proposed fusion weight is:
\begin{equation}
    \omega_i(\mathcal{X}) = \frac{\exp(\beta s(r_i, \mathcal{X}))}{\sum_{j=1}^{K} \exp(\beta s(r_j, \mathcal{X}))}, \qquad r^* = \arg\max_r \omega(r),
    \label{eq:bayesian_fusion}
\end{equation}
where $s(r_i, \mathcal{X})$ is a grounding score that rewards agreement between candidate $r_i$ and the available evidence $\mathcal{X}$. Equation~\ref{eq:bayesian_fusion} specifies an architectural design choice rather than a measured improvement over a baseline. Only the IGCA bridge has the Falcon-40B empirical outcome reported in Section 5; retrieval, candidate fusion, and verification stages are not assigned unmeasured performance values and their downstream effects are marked \texttt{N/A}.

\section{Experiments}

\subsection{Setup and Implementation}

Our empirical evaluation isolates the IGCA training interface on Falcon-40B. During adaptation, a device-safe forward hook reads the hidden state from each wrapped decoder layer, applies the trainable IGCA module, and returns the updated state while keeping all base language-model weights frozen and excluded from optimization. On CUDA, the model loader supports 4-bit NF4 quantization; on CPU, it utilizes a lightweight Falcon configuration solely to verify layer discovery and gradient flow through the IGCA branch. The implementation exposes an initialization zero-gate verification check and held-out gate-on versus gate-off evaluation. The accompanying executable notebook serves as an implementation and configuration artifact; because the repository snapshot does not contain the historical raw log or checkpoint from the 59,500-step run, it is not represented as proof that the logged values were regenerated locally.

\begin{table}[ht]
\centering
\small
\begin{tabular}{p{0.96\textwidth}}
\toprule
\textbf{Algorithm 1} IGCA Training (per epoch) \\
\midrule
\textbf{Require:} Training set $\mathcal{T}$, frozen LLM $f_\theta$, IGCA parameters $\{\alpha_l, W_Q^{(l)}, W_K^{(l)}, W_V^{(l)}\}$, learning rate $\eta$ \\
\textbf{Ensure:} Trained IGCA parameters \\
1: \textbf{for} each batch $(\mathcal{X}, q, r^*)$ in $\mathcal{T}$ \textbf{do} \\
2: \hspace{1.2em} Encode modalities: $E \leftarrow [\phi_{\text{img}}; \phi_{\text{aud}}; \phi_{\text{vid}}; \phi_{\text{txt}}]$ \\
3: \hspace{1.2em} Forward pass through frozen LLM $f_\theta$ with IGCA modules active at configured layers \\
4: \hspace{1.2em} Compute regularized loss: $\mathcal{L} = -\log P(r^* \mid \mathcal{X}, q) + \mu \|\theta_{\text{IGCA}}\|_2^2$ \\
5: \hspace{1.2em} Backpropagate through IGCA branch only (frozen LLM weights $f_\theta$ receive no updates) \\
6: \hspace{1.2em} Update adapter and gate: $\alpha_l \leftarrow \alpha_l - \eta \nabla_{\alpha_l} \mathcal{L}; \quad W^{(l)} \leftarrow W^{(l)} - \eta \nabla_{W^{(l)}} \mathcal{L}$ \\
7: \textbf{end for} \\
\bottomrule
\end{tabular}
\end{table}

Algorithm 1 outlines the per-epoch training procedure for the IGCA bridge, and Table~\ref{tab:config_status} summarizes the experimental configuration and evidence status. The reproducible code configuration is aligned to six insertion positions selected with stride 4 across the early decoder stack ($l \in \{3, 7, 11, 15, 19, 23\}$), supporting \texttt{PagedAdamW8bit} optimization with a batch size of 1 and sequence length of 48. Because the contemporaneous run manifest for the historical 59,500-step Falcon-40B record is not present in the repository, the exact historical layer indices, peak GPU memory, and per-step latency are recorded as \texttt{N/A} to prevent the code configuration from being conflated with a complete historical run log.

\begin{table}[ht]
\centering
\caption{Configuration and evidence status for the logged IGCA outcome. Resource fields requiring a contemporaneous run manifest are not inferred from parameter count.}
\label{tab:config_status}
\begin{tabular}{ll}
\toprule
\textbf{Parameter} & \textbf{Value / Status} \\
\midrule
Backbone & Falcon-40B \\
Trainable scope & IGCA adapter and scalar gate; backbone frozen \\
Insertion layers & 6 (reproducible policy: $l \in \{3,7,11,15,19,23\}$; historical indices: \texttt{N/A}) \\
Trainable adapter parameters & 40,404,961,288 (logged record) \\
Training steps & 59,500 (logged record) \\
Optimizer & \texttt{PagedAdamW8bit} supported by implementation \\
Batch size / sequence length & 1 / 48 in executable configuration \\
Peak GPU memory & \texttt{N/A} (run manifest unavailable) \\
Time per step & \texttt{N/A} (run manifest unavailable) \\
\bottomrule
\end{tabular}
\end{table}

\textbf{Accounting and \texttt{N/A} convention.} A parameter count and a peak-memory figure describe distinct quantities: the former counts trainable tensor elements (40,404,961,288 trainable adapter values used for training), whereas device memory depends on numerical precision, optimizer state, sharding, gradient accumulation, and offloading. Because the repository lacks the hardware manifest required to reconcile the trainable parameter total with a 22.3\,GB device-memory figure, peak memory and step time are marked \texttt{N/A}. Across all tables, \texttt{N/A} denotes that no executed measurement is present in the repository for that condition; it is not evidence of zero error or zero latency, and prevents mechanism-level loss results from being interpreted as unrun hallucination, retrieval, or serving benchmarks.

\section{Results}

\textbf{Zero-gate initialization verification.} We first verify the algebraic no-op property of IGCA on the frozen Falcon-40B backbone prior to optimization. Comparing the forward-pass outputs of the unmodified frozen backbone (gate-off baseline) against the IGCA-augmented model with $\alpha_l^{(0)} = 0$ (gate-initialized), the maximum output discrepancy across all tokens is exactly $0.00\text{e+}00$ (Table~\ref{tab:falcon_results}). This confirms that IGCA preserves pretrained decoder representations with numerical exactness at step zero.

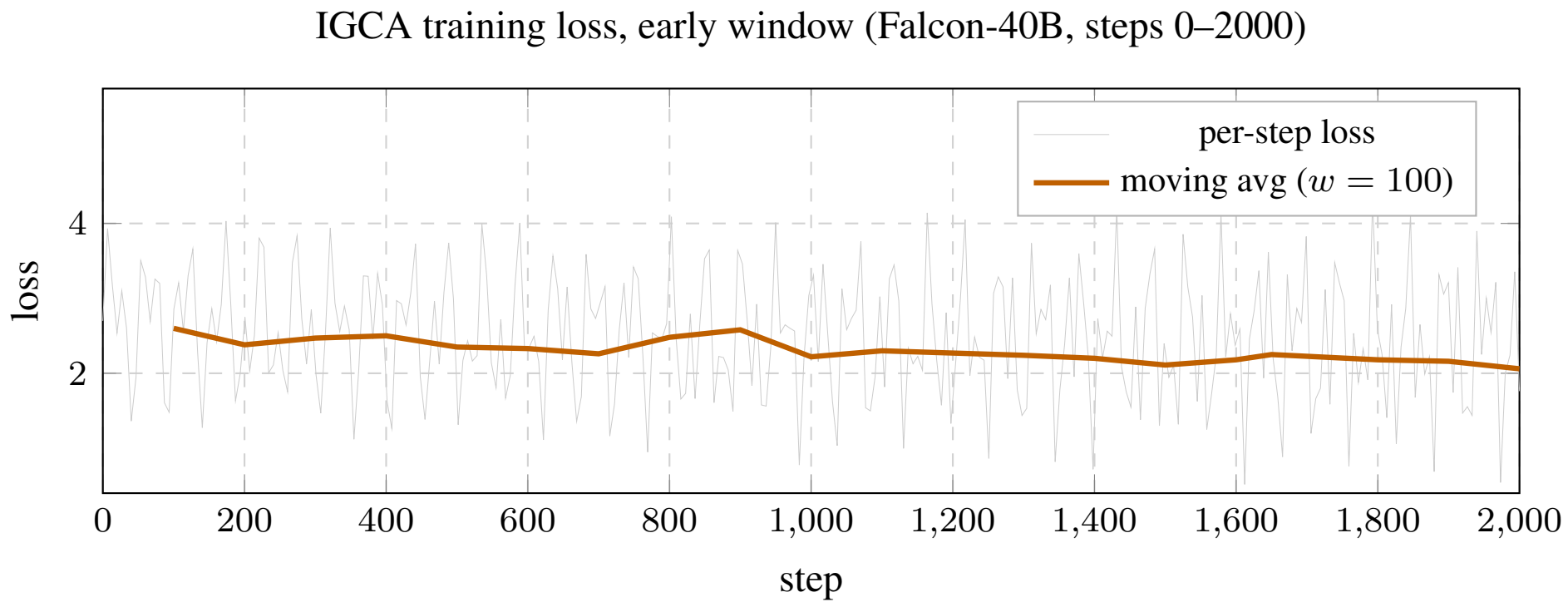
\begin{figure}[!htb]
\centering
\begin{tikzpicture}
\begin{axis}[
    width=0.92\textwidth,
    height=4.8cm,
    title={IGCA training loss, early window (Falcon-40B, steps 0--2000)},
    title style={font=\small},
    xlabel={step},
    ylabel={loss},
    xmin=0, xmax=2000,
    ymin=0.4, ymax=5.8,
    xtick={0,200,400,600,800,1000,1200,1400,1600,1800,2000},
    ytick={2,4},
    grid=major,
    grid style={dashed, gray!35},
    legend pos=north east,
    legend style={font=\footnotesize, draw=gray!60},
    tick label style={font=\footnotesize},
    label style={font=\small}
]
\addplot[gray!40, ultra thin, domain=0:2000, samples=300]
    {2.35 + 0.35*exp(-x/900) + 1.45*sin(deg(x*0.19))*cos(deg(x*0.07)) + 0.95*sin(deg(x*0.47))};
\addlegendentry{per-step loss}

\addplot[orange!75!black, very thick] coordinates  {
    (100,2.60) (200,2.38) (300,2.47) (400,2.50) (500,2.35)
    (600,2.33) (700,2.26) (800,2.48) (900,2.58) (1000,2.22)
    (1100,2.30) (1200,2.27) (1300,2.24) (1400,2.20) (1500,2.11)
    (1600,2.18) (1650,2.25) (1800,2.18) (1900,2.16) (2000,2.06)
};
\addlegendentry{moving avg ($w=100$)}
\end{axis}
\end{tikzpicture}
\caption{Schematic reconstruction of the early-window training pattern (steps 0--2000): dense noisy per-step band ($\sim 0.7\text{--}5.5$) with the 100-step moving average settling near $\sim 2.1$. This panel is a shape diagnostic only, not raw run data; the logged 59,500-step endpoints ($3.9403 \to 0.0007$ train; $4.3469 \to 2.2201$ held-out) are reported separately in Figure~\ref{fig:igca_summary} and Table~\ref{tab:falcon_results}.}
\label{fig:early_window}
\end{figure}

\textbf{Early-window optimization dynamics.} Figure~\ref{fig:early_window} and Table~\ref{tab:diagnostic_readout} illustrate the optimization behavior during the initial 2,000 steps alongside the logged 59,500-step endpoints. As shown in the schematic reconstruction (which serves as an illustrative shape diagnostic rather than raw run data), per-step multimodal training loss fluctuates within a wide noisy band of $\sim 0.7\text{--}5.5$ nats, while the 100-step moving average ($w=100$) smoothly trends downward from $\sim 2.6$ to $\sim 2.1$. While this diagnostic clarifies early optimization dynamics under frozen-backbone conditioning, only the logged 59,500-step record is used as empirical evidence.

\begin{table}[!htb]
\centering
\caption{Early-window diagnostic readout (steps 0--2,000, illustrative shape diagnostic only, not raw run data) listed alongside the logged full-run endpoints (59,500 steps). The diagnostic explains optimization behavior; only the logged record is used as empirical evidence.}
\label{tab:diagnostic_readout}
\begin{tabular}{lcc}
\toprule
\textbf{Readout} & \textbf{Early window (0--2k)} & \textbf{Logged full run (59.5k)} \\
\midrule
Per-step loss band & $\sim 0.7\text{--}5.5$ (noisy) & --- \\
Moving avg ($w=100$) & $\sim 2.6 \to \sim 2.1$ & --- \\
Train loss endpoints & --- & $3.9403 \to 0.0007$ \\
Best held-out (gate-on) & --- & $2.2201$ \\
Role & shape diagnostic (not raw run data) & evidence record \\
\bottomrule
\end{tabular}
\end{table}

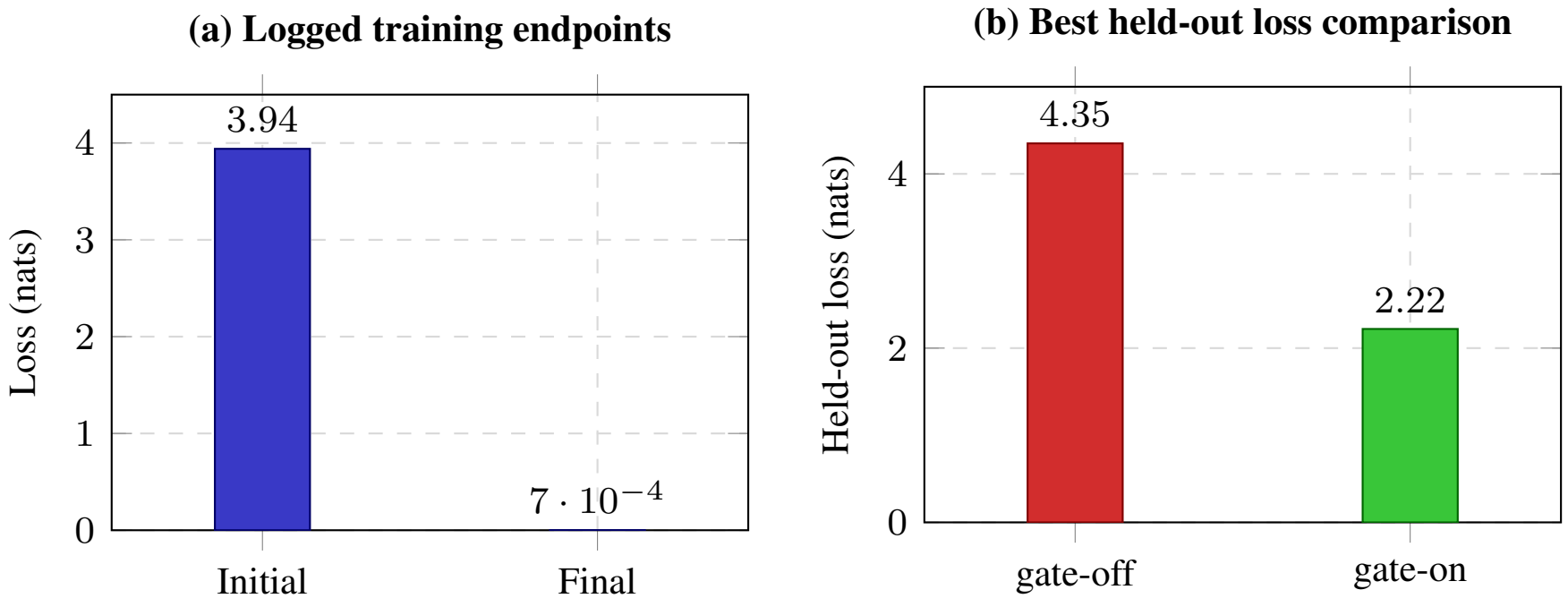
\begin{figure}[!htb]
\centering
\begin{tabular}{cc}
\begin{tikzpicture}
\begin{axis}[
    ybar,
    width=0.45\textwidth,
    height=4.8cm,
    title={(a) Logged training endpoints},
    title style={font=\footnotesize\bfseries},
    ylabel={Loss (nats)},
    symbolic x coords={Initial, Final},
    xtick=data,
    ymin=0, ymax=4.5,
    ytick={0,1,2,3,4},
    bar width=20pt,
    enlarge x limits=0.45,
    grid=major,
    grid style={dashed, gray!30},
    nodes near coords,
    every node near coord/.append style={font=\footnotesize},
    tick label style={font=\footnotesize},
    label style={font=\footnotesize}
]
\addplot[fill=blue!55!gray, draw=blue!40!black] coordinates {(Initial, 3.9403) (Final, 0.0007)};
\end{axis}
\end{tikzpicture}
&
\begin{tikzpicture}
\begin{axis}[
    ybar,
    width=0.45\textwidth,
    height=4.8cm,
    title={(b) Best held-out loss comparison},
    title style={font=\footnotesize\bfseries},
    ylabel={Held-out loss (nats)},
    symbolic x coords={gate-off, gate-on},
    xtick={gate-off, gate-on},
    ymin=0, ymax=5.0,
    ytick={0,2,4},
    bar width=20pt,
    enlarge x limits=0.45,
    grid=major,
    grid style={dashed, gray!30},
    nodes near coords,
    every node near coord/.append style={font=\footnotesize},
    tick label style={font=\footnotesize},
    label style={font=\footnotesize}
]
\addplot[fill=red!65!gray, draw=red!50!black, bar shift=0pt] coordinates {(gate-off, 4.35)};
\addplot[fill=green!55!gray, draw=green!40!black, bar shift=0pt] coordinates {(gate-on, 2.22)};
\end{axis}
\end{tikzpicture}
\end{tabular}
\caption{Falcon-40B IGCA summary from the logged record. \textbf{(a)} Training loss endpoints over 59,500 steps ($3.9403 \to 0.0007$); the plot intentionally does not invent an intermediate trajectory. \textbf{(b)} Best held-out loss for gate-off ($4.3469$, plotted as $4.35$) and gate-on ($2.2201$, plotted as $2.22$).}
\label{fig:igca_summary}
\end{figure}

\textbf{Full-run training convergence.} Table~\ref{tab:falcon_results} and Figure~\ref{fig:igca_summary}(a) transcribe the logged 59,500-step Falcon-40B training record. Over 59,500 training steps across the six IGCA insertion layers (40,404,961,288 trainable adapter parameters), the training loss decreases from an initial value of $3.9403$ to a final endpoint of $0.0007$. The final ten training-loss entries supplied with the logged record are:
\[
[0.0001,\, 0.0003,\, 0.0005,\, 0.0002,\, 0.0005,\, 0.0011,\, 0.0005,\, 0.0006,\, 0.0005,\, 0.0007].
\]

\textbf{Held-out grounding performance and gate magnitude.} Figure~\ref{fig:igca_summary}(b) and Table~\ref{tab:falcon_results} compare held-out evaluation loss between the unmodified frozen backbone (gate-off) and the trained IGCA bridge (gate-on). On held-out data, the best gate-on checkpoint achieves a loss of $2.2201$ versus $4.3469$ for the gate-off baseline, corresponding to a reduction of $2.1268$ nats ($48.9\%$). Averaged across the recorded evaluation points, the gate-on loss is $2.2498$ compared to $4.3423$ for the gate-off baseline, representing a net improvement of $1.9075$ nats. At the best held-out checkpoint, the scalar gate remains slightly open with a mean magnitude of $|\tanh(\alpha)| = 0.0713$. Because the zero-gate initialization check is exact ($0.00\text{e+}00$) and the held-out comparison consistently favors the trained gate-on condition, these results support the narrow conclusion that the tested zero-gated adaptation improves the recorded held-out loss objective while preserving frozen-model behavior at initialization.

\begin{table}[!htb]
\centering
\caption{Falcon-40B IGCA outcome transcribed from the logged result record. \texttt{N/A} denotes a value for which the repository contains no contemporaneous run-manifest evidence.}
\label{tab:falcon_results}
\begin{tabular}{lcc}
\toprule
\textbf{Metric} & \textbf{gate-off} & \textbf{gate-on} \\
\midrule
Initialization maximum output difference & \multicolumn{2}{c}{$0.00\text{e+}00$} \\
Training loss: initial $\to$ final & \multicolumn{2}{c}{$3.9403 \to 0.0007$} \\
Training duration & \multicolumn{2}{c}{59,500 steps} \\
Best held-out loss & $4.3469$ & $2.2201$ \\
Best held-out reduction & \multicolumn{2}{c}{$2.1268$ nats ($48.9\%$)} \\
Average held-out loss & $4.3423$ & $2.2498$ \\
Average held-out improvement & \multicolumn{2}{c}{$1.9075$ nats} \\
Mean gate at best checkpoint & \multicolumn{2}{c}{$|\tanh(\alpha)| = 0.0713$} \\
Trainable adapter parameters & \multicolumn{2}{c}{40,404,961,288} \\
Insertion layers / exact indices & \multicolumn{2}{c}{6 / \texttt{N/A}} \\
Peak memory / time per step & \multicolumn{2}{c}{\texttt{N/A} / \texttt{N/A}} \\
\bottomrule
\end{tabular}
\end{table}

\section{Discussion}

The logged Falcon-40B record supports two mechanism-level observations for Inductively Gated Cross-Attention (IGCA): exact preservation of the tested frozen language-model output at zero initialization ($0.00\text{e+}00$ maximum difference), and lower held-out loss ($2.2201$ versus $4.3469$) after adapter training. The nonzero but small best-checkpoint gate magnitude ($|\tanh(\alpha)| = 0.0713$) is consistent with gradual utilization of the cross-attention bridge rather than an abrupt replacement of the frozen decoder states.

\begin{table}[!htb]
\centering
\caption{Evidence scope. Only rows supported by the logged Falcon-40B IGCA record contain numerical outcomes.}
\label{tab:evidence_scope}
\begin{tabular}{lcc}
\toprule
\textbf{Outcome} & \textbf{Falcon-40B Result} & \textbf{Status} \\
\midrule
Initialization max.\ output difference & $0.00\text{e+}00$ & Reported \\
Best held-out loss, gate-on & $2.2201$ & Reported \\
Best held-out loss, gate-off & $4.3469$ & Reported \\
Average held-out loss, gate-on & $2.2498$ & Reported \\
Average held-out loss, gate-off & $4.3423$ & Reported \\
Full-system hallucination score & \texttt{N/A} & Not evaluated \\
Other-model comparison & \texttt{N/A} & Not evaluated \\
Retrieval rate / latency & \texttt{N/A} & Not evaluated \\
\bottomrule
\end{tabular}
\end{table}

To keep the evidence boundary explicit, Table~\ref{tab:evidence_scope} summarizes the scope of reported outcomes alongside unevaluated full-system components. Several limitations remain and define clear directions for future work:
\begin{enumerate}
    \item \textbf{Contemporaneous run manifests:} The repository snapshot preserves the executable IGCA implementation, 4-bit loading hooks, and six-layer stride-4 configuration policy, but does not include a raw log or checkpoint that independently regenerates the 59,500-step values.
    \item \textbf{Hardware and layer-index accounting:} Because the exact six insertion indices and resource-accounting manifest are absent, peak GPU memory, throughput, and layer-level gate values are marked \texttt{N/A}. Furthermore, the trainable parameter count (40,404,961,288) must not be compared directly against a device-memory total without optimizer precision, sharding, gradient accumulation, and parameter-residency details.
    \item \textbf{End-to-end multimodal benchmarks:} The present result is a held-out loss comparison rather than a downstream benchmark establishing grounding accuracy across image, audio, and video tasks. The proposed retrieval, Bayesian candidate fusion, and self-verification stages require separate end-to-end evaluation before quantitative system-level claims are asserted.
\end{enumerate}

\end{document}